# AnalogVNN: A Fully Modular Framework for Modeling and Optimizing Photonic Neural Networks

Vivswan Shah, and Nathan Youngblood

*Department of Electrical and Computer Engineering, University of Pittsburgh, Pittsburgh, PA 15261*
*Corresponding author: vivswanshah@pitt.edu and nathan.youngblood@pitt.edu*

**Abstract:** In this paper, we present AnalogVNN, a simulation framework built on PyTorch which can simulate the effects of optoelectronic noise, limited precision, and signal normalization present in photonic neural network accelerators. We use this framework to train and optimize linear and convolutional neural networks with up to 9 layers and ~1.7 million parameters, while gaining insights into how normalization, activation function, reduced precision, and noise influence accuracy in analog photonic neural networks. By following the same layer structure design present in PyTorch, the AnalogVNN framework allows users to convert most digital neural network models to their analog counterparts with just a few lines of code, taking full advantage of the open-source optimization, deep learning, and GPU acceleration libraries available through PyTorch.

## I. Introduction

In the past decade, there has been exponential growth in the size, complexity, efficiency, and robustness of deep neural networks [1], [2]. However, the computing resources needed to fuel this continuous advancement in machine learning have also grown exponentially - much faster than the performance and efficiency improvements of the hardware used to train networks [2] or perform inference [3]. The computation of DNNs is comprised of approximately 90% linear operations (matrix-vector multiplications or convolutions) and 10% simple nonlinear operations (e.g., sigmoid, ReLU, tanh, etc.) [4]. These linear operations can be processed with high efficiency and low latency using analog computational schemes by leveraging parallelized circuits in the analog domain. For example, a single READ operation on a memristor array can perform an entire matrix-vector multiplication in a single clock cycle [5]–[7]. So, by performing the computation in the analog domain, we can significantly reduce the training and inference times of neural networks [8]. One particularly attractive approach is photonic analog computing which promises ultra-low latency, high energy efficiency, and unparalleled data throughput [9]–[14]. This is possible due to the enhanced modulation speeds of optical waveguides and fibers over their electronic counterparts which suffer from resistive and capacitive losses [15]. Thus, by using photonics, the large-scale linear operations of neural networks can be performed efficiently at high modulation speeds and ultra-low latencies, potentially leading to networks with high throughput and real-time processing [16].

Translating neural network models directly from the digital domain to any other domain (the photonic analog domain in our case) without any changes to its structure or hyperparameters will cause a significant reduction in the accuracy and generalizability of the model. Neural networks are explicitly trained to operate in the environment they were trained in. Simply translating the weight or structure of a network to a new environment with reduced precision or increased noise is as problematic as it is the biological counterpart of a head transplant. By carefully designing neural networks based on their computing environment (digital, analog, or physical), a significant improvement in the model's generalizability, accuracy, and ability to learn the crucial features of the dataset can be seen [17]. For analog DNN accelerators, this can be done by implementing models on-chip for different hyperparameter combinations and testing each of them, but this is time-consuming and costly to do. To overcome this problem, we can simulate computation in the analog domain and test against various hyperparameters virtually. It has already been shown that even a crude simulation of the target domain can overcome much of the accuracy and generalizability problems [17], [18].

We have developed the Analog Virtual Neural Network (AnalogVNN) framework [19] to do exactly this. Distinct from other approaches which typically focus on modeling the physical response of the analog hardware in question [8], [17], [18], [20]–[22], we have chosen to abstract the physical properties of the analog hardware and instead model the effects an ensemble of analog computing elements at a higher level (i.e., normalization, limited precision, stochastic rounding, and additive noise). This approach greatly simplifies the translation of digital neural network models to the analog domain, while minimizing the additional computational overhead required to model analog hardware as illustrated in **Figure 1a**. We have built AnalogVNN on PyTorch [23] to easily simulate the effects of optoelectronic noise, limited precision, and signal normalization present in all photonic analog hardware. While we have designed the AnalogVNN framework with photonic hardware in mind (e.g., coherent [11], [24], electro-absorptive [25], phase-change [26], microring resonator [27], and dispersive fiber-based architectures [12] as illustrated in **Figure 1b**), the generality of our approach allows all researchers to easily extend our work to other analog neural networks, such as those based on electronic, magnetic, or spintronic hardware [8], [9], [28]–[30].

The repository for AnalogVNN is available at *https://analogvnn.github.io*
Sample code: *https://analogvnn.github.io/sample_code*

## II. The Analog Virtual Neural Network (AnalogVNN) Methodology

The photonic analog domain differs from the digital domain in two major ways. First, one has to account for continuous variability due to added noise from physical processes and second, the precision is typically limited by photon shot noise to 8-bits or less for the optical powers and modulation speeds of interest [31]. In the case of photonic weights, physical processes such as thermal drift in microring resonators or stochastic effects in the programming of phase-change photonic memory, introduce stochastic noise to the weight matrix. Photonic analog inputs, on the other hand, have been limited to even lower precision in practice (e.g., 4-bit for PAM-16 modulators). As high-speed optical modulators have primarily targeted telecommunication applications, they are typically designed to generate a limited set of optical amplitudes which minimize bit error rates from optoelectronic noise and timing jitter. These characteristics can be abstracted and simulated by adding intermediate layers which intentionally introduce noise and reduce precision (using Noise and Reduce Precision layers respectively) to a digital model for the linear analog system (**Figure 1c-d**). In this way, the digital models are able to efficiently imitate the analog environment, and exploration of analog hyperparameters can be achieved more effectively. The optimization of the network can therefore be more efficient, and we can begin to identify hyperparameters that improve the model's generalizability, accuracy, learning rate, and the ability for the network to learn the crucial features of the dataset [32]–[36].

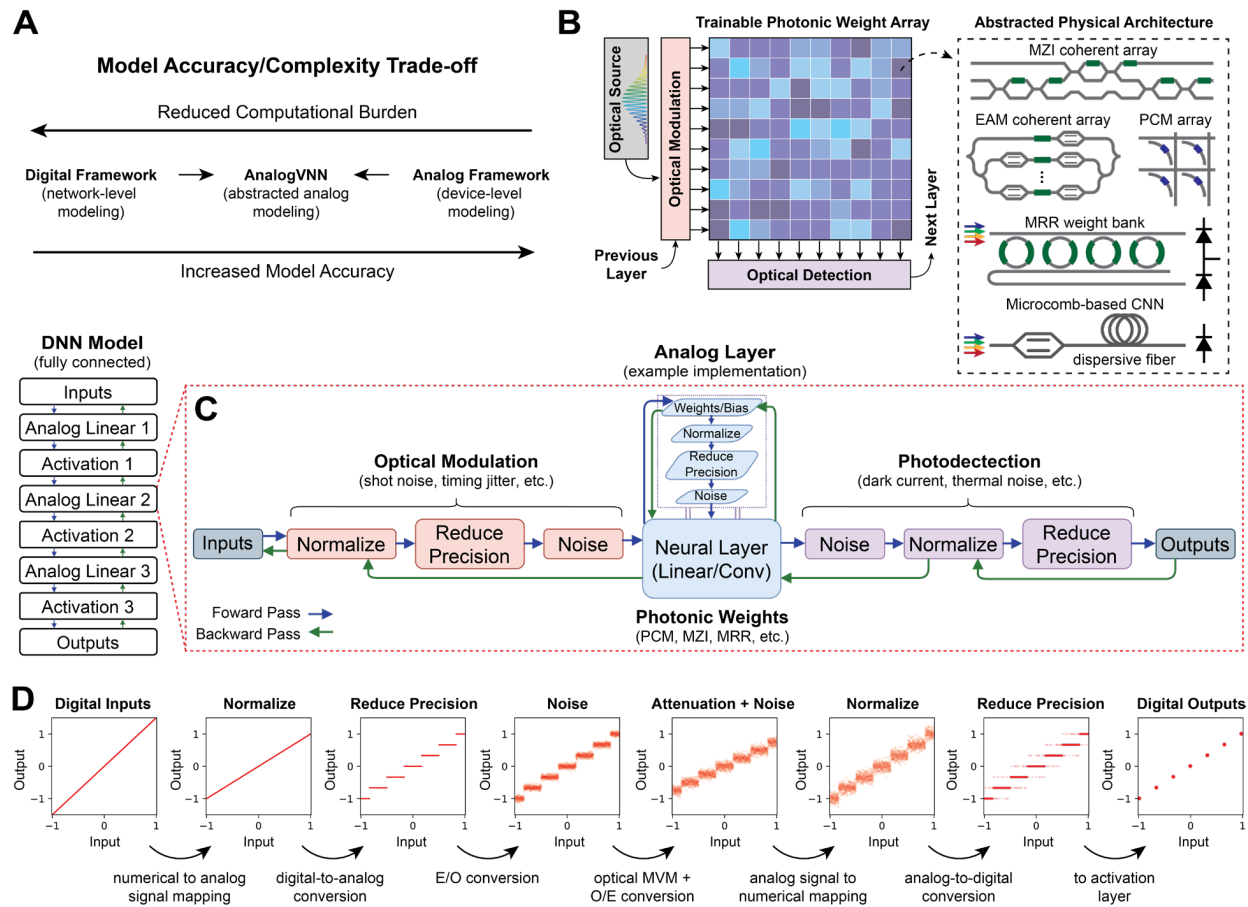

**Figure 1: Overview of the AnalogVNN framework. a)** Trade-off between computational burden and accuracy of model used to simulate analog AI hardware. AnalogVNN bridges the gap between fully digital and fully analog frameworks. **b)** Conceptual illustration of the three main abstraction layers within AnalogVNN: optical modulation (input vector), trainable photonic weights (matrix operation), and optical detection (output vector). The physical implementation of the photonic hardware is abstracted to normalization, noise, and reduced precision layers, allowing our framework to be applied to various photonic architectures. **c)** Overview of a 3-layer linear model with three Analog Linear and Activation layers. The inner working of an Analog Linear layer is shown in the red dotted box with its corresponding on-chip photonic implementation. Blue and Green arrows represent forward and backward pass respectively. **d)** Illustration of the optoelectronic analog effects modeled during a forward pass through a linear analog layer.

*Contributions*

Simulating an analog neural network with PyTorch or TensorFlow requires the aforementioned additional features which are not present in current frameworks. The first requirement is the ability to create parameterized weights, biases, and layers without affecting the gradient flow graph of the network. As an example, one should be able to add normalization, precision, and noise to weights, biases, or between layers and still retain the ability to calculate as if these additional layers were not present. Second, training analog network models may require a new analog optimizer to train efficiently. These optimizers can be created by combining the properties of the analog domain with those already well-established digital optimizers (like Adam, SDG, etc.). Nandakumar et al. 2020 [8] show an example of this in which a new Reduce Precision Optimizer was used to train an on-chip network faster and more efficiently. Third, as stated earlier, by following the same layer structure design present in PyTorch, AnalogVNN allows users to convert most digital neural network models to their analog counterparts with just a few lines of code. When comparing PyTorch sample code [37] from its tutorial to AnalogVNN sample code [38], the only differences which are specific to AnalogVNN can be found in *add layer* function (which adds Reduce Precision, Noise and Normalization layers) and *PseudoParameter* (which converts digital parameters into analog parameters), hence adding only 12 new lines code unique to AnalogVNN. Because we have built AnalogVNN with PyTorch [23] modularity and compatibility in mind, this provides ease of use and same-day access (i.e., zero-day access) to future new PyTorch features.

This combination of features and options provides a robust and customizable environment for researchers to design, simulate, and test arbitrary photonic or analog neural network models based on other similar hardware. We first use AnalogVNN to design and optimize hyperparameters in small 3- to 6-layer photonic image classification models. We then show the generality of our conclusions from these smaller models by optimizing the larger and more complex 9-layer CIFAR-10 convolutional neural network (CNN) model by P. Kaur for the photonic analog domain [8], [39]. The main features of AnalogVNN, which we used in this paper are the ability to control the gradient flow graph (e.g., skipping noise layer and reduce precision layer during backpropagation, see **Figure 1c**) and the introduction of noise, reduce precision, and normalization to model parameters (e.g., the weights and biases of the network) as shown in **Figure 1d**. Finally, after testing and training, the final analog neural network with optimized hyperparameters can be transferred to a photonic chip for on-chip optimization.

## III. Analog Layer Design Approach

To design a layer that can simulate an analog system such as a photonic processor, we look at the major factors which make a photonic network different from a digital network. Namely, photonic and digital networks differ in their inputs, outputs, and storage units (weights and biases). In digital systems, inputs can have very high precision ($2^{16}$ discrete levels or 16-bit floating point is used by PyTorch and TensorFlow [23], [40], [41]) and very low noise (negligible noise due to digital operation), while in photonic systems inputs are generated by optical sources (typically lasers) and are limited by the physical output characteristics of these optical sources. Relative intensity noise (RIN) and modulation frequency place a fundamental upper bound on the precision of the optical inputs. For example, to encode an 8-bit analog signal with a low noise laser source (RIN = −165 dB/Hz), the modulation frequency will be limited to around 4 GHz or less regardless of the optical power [31]. In addition to relative intensity noise, other noise sources such as optical shot noise from the laser, thermal noise from the analog driving circuitry, and timing jitter, can contaminate the analog signal and limit the maximum precision of the inputs [42]. So, while digital inputs can take on a very large range of values due to their high precision, this is not true for analog systems. Analog inputs also typically operate on a system of regular divisions across a relative scale rather than the binary representation of maximal values across many bits like in digital systems. For example, the maximum power of the laser is typically mapped to encode a normalized value between 0 and 1, while the phase of the optical input can be used to encode negative values in coherent architectures. Additionally, analog systems driven by digital inputs are made pseudo-analog to encode discrete digital values in the presence of noise (e.g., as in PAM-4, PAM-16, or 64-QAM modulation schemes). To virtualize these effects illustrated in **Figure 2a**, we first normalize the input signals, divide them to a certain precision, and then add noise. It is important for noise to be added after digitization since it would otherwise be removed by the digitization process. Therefore, the differences between inputs in the digital and photonics domains can be simulated using: 1) Normalization, 2) Reduce Precision, and 3) Noise layers to represent the analog response to digital inputs (see **Figure 2b**). Details on the mathematical implementation of each of these layers can be found in the Appendix.

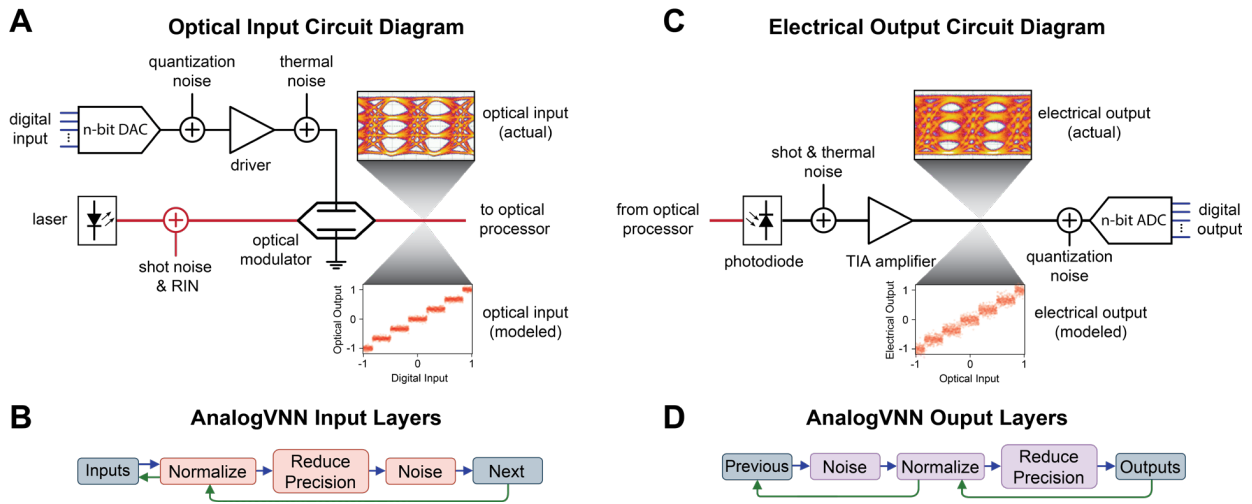

Figure 2: Modeling approach for analog input and output layers. a) Example optoelectronic circuit schematic and b) AnalogVNN implementation for encoding analog inputs to the optical processing unit. Blue, black, and red circuit connections represent digital, electrical, and optical interconnects respectively. Various sources of noise can be modeled by adding multiple noise layers in series. c) Typical electrical circuit diagram and d) AnalogVNN implementation used to convert processed optical signals back to the digital domain. During the hyperparameter exploration in the following sections, a single noise Gaussian noise layer was used in both the optical input and electrical output layers for simplicity.

While a laser is used to create the photonic analog inputs, photodetectors are typically used to measure the analog output signals of the photonic processor (illustrated in **Figure 2c**). Similar to lasers, photodetectors and their subsequent gain stages are analog components that exhibit limited precision due to physical effects such as shot noise due to dark current, read noise due to amplifier circuitry, Johnson noise due to thermal effects, and nonlinear response at high input powers due to saturation [43]. Hence a similar layer structure as the inputs can be used for the outputs but a different order is needed to model the physical processes: 1) Noise, 2) Normalization, and 3) Reduce Precision layers as shown in **Figure 2d**.

The final component to consider for data processing in an analog layer is the matrix-vector operations performed in the linear and convolutional layers of neural networks (**Figure 3**). The neural layer is comprised of two components: 1) the data associated with the layer-like weights and biases, and 2) the compute operation done within the layer. First, the data associated with the layer such as weights and biases can be directly encoded with the help of on-chip modulators (e.g., MZI, microring, or electro-absorption modulators [11], [13], [27], [44]–[46]), fiber-based programmable filters [12], free space components (e.g., spatial light modulators [47], [48]), or phase change materials [10], [49]–[51]. In the absence of on-chip gain, these components can encode numbers in the range of $[-1, 1]$ since for incoherent architectures, balanced photodetectors can be used for differential detection [27]. **Figure 3a-b** illustrate two examples of such photonic matrix-vector multiplier implementations where either microring modulators or nonvolatile phase-change materials are used to encode the elements of matrix $W$. In the case of on-chip encoded data, there may be a separate limitation on the precision and a minimal amount of noise or repeatable noise. Consequentially, the value of trainable weights must be normalized with respect to their maximum transmission before reducing their precision, resulting in the combination of Normalization and Reduce Precision layers in that order (see **Figure 3c**). However, in the case of weights being encoded in the optical signal (e.g., time-multiplexed architectures [52]–[54]), the same characteristics as for the analog input layer will apply.

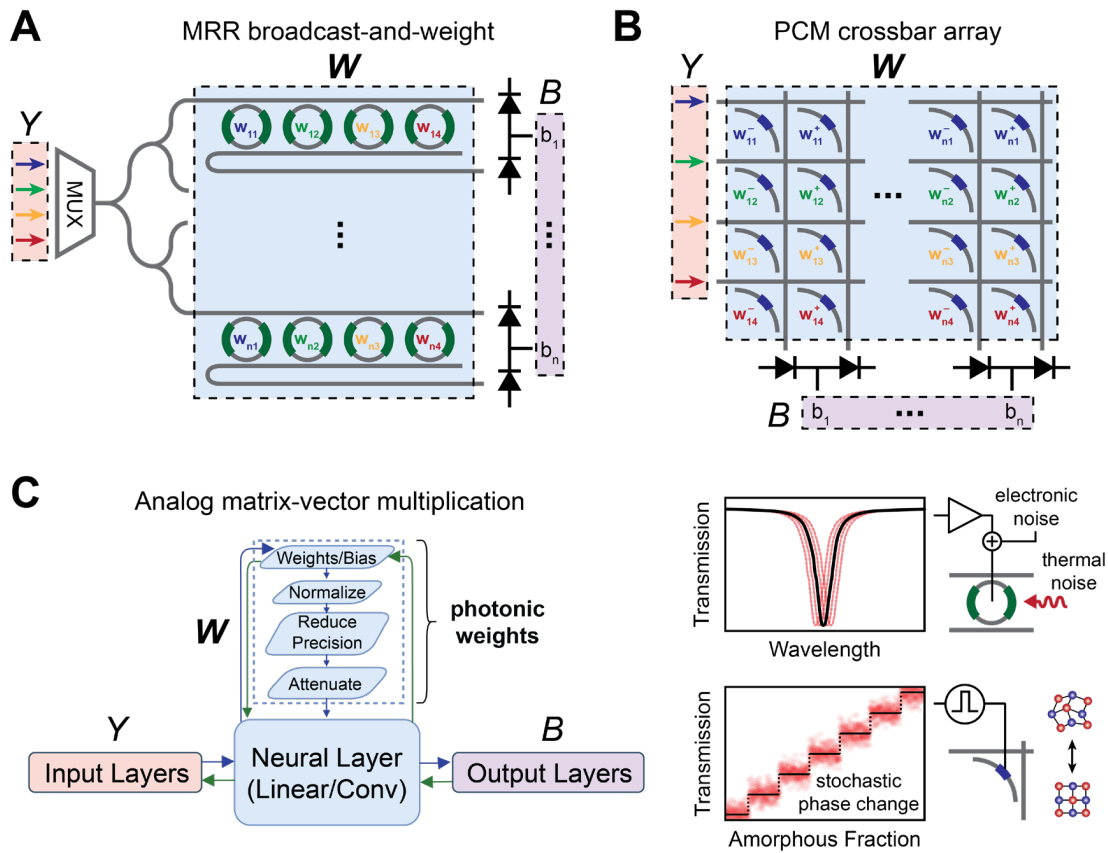

**Figure 3: Modeling approach for the linear and convolutional layers. a)** Example implementations of a photonic matrix-vector multiplier using an array of **a)** microring resonator (MRR) modulators [27] and **b)** phase-change photonic memory [27]. **c)** AnalogVNN modeling approach for implementing linear and convolutional layers (right). Example sources of noise, stochasticity, and reduced precision in photonic weights for the architectures illustrated in **a)** and **b)** (left).

Second, photonic circuits are intrinsically lossy and imperfect due to fabrication limitations and material absorption. Thus, the resulting system exhibits intensity loss in each part of the compute operation such as during addition, multiplication, or any other functional operations. Additionally, noise in the optical weights can arise due to random fluctuations from thermal drift [55] and noise from the analog driving circuitry. Other non-dynamic, stochastic processes which occur during programming (e.g., crystallization in PCMs [56]) can be accounted for in the Reduce Precision layer mentioned previously (see the "Stochastic Reduce Precision" layer definition in the Appendix). Parameterizing compute operations in these analog weight layers is computationally very expensive (more details in the following section), so to approximate intensity loss and noise in each part of the operation, we assume that all operations are performed first without error and then add the intensity loss and noise before sending the result to the following analog output layers as illustrated in **Figure 3c**. However, since the analog output simulation already contains a noise layer, it is possible to combine the noise from the weights with that of the analog output layer for uncorrelated noise of the same distribution. Likewise, any equivalently proportional intensity loss through the photonic weights will be removed by the analog output normalization layer, though the non-normalized portion of the signal with respect to the uniform gaussian noise [signal-to-noise ratio] will require that the added noise in the output layer is scaled accordingly. Hence, we can reduce the simulation time of our analog network by removing these duplicate layers when applicable.

### *Training Optimization for AnalogVNN Models*

Here, we briefly note that if one directly attempts to simulate analog neural networks in PyTorch without updating the optimizer and parameters as is done in this paper, PyTorch will be unable to calculate gradients due the addition of reduce precision and noise layers, whose values are discontinuous. As the gradient can no longer be calculated from the discontinuities, no training can occur. Therefore, it is necessary to upgrade the parameters by using AnalogVNN's *PseudoParameter* classes, to enable training in PyTorch. In short, the process of converting digital neural networks to analog neural networks in AnalogVNN requires the following steps to be performed: First, convert the model by adding analog layers between all of the digital layers to be emulated within the model. Second, convert the weights and biases by running the *PseudoParameter* class on the initial digital model to generate a new set of analog parameters (weight and biases). Third, convert the digital optimizer to work with analog parameters so that the training itself mimics an analog system is automatically done just by using *PseudoParameter* to convert digital parameter into analog parameter. Lastly, the training can be executed per the typical design flow and using the same instructions as those otherwise required in PyTorch due to the optimizers in AnalogVNN providing all the similar functions as PyTorch for training. For further instruction, please refer to the AnalogVNN Documentation [19].

## IV. Results and Discussion

*Hyperparameter Exploration for Small-Scale Analog Neural Networks (1- to 6-layer)*

To explore the effects that various hyperparameters on analog photonic neural networks, we tested 1- to 6- layers image classification models with over 570,000 different hyperparameter combinations each using our AnalogVNN framework. This large exploration of hyperparameter space is made possible by our approach which simplifies the modeling of analog domain effects and makes use of PyTorch's optimized GPU libraries during training. We found that using AnalogVNN results in less than 10% longer training times compared to equivalent digital models without added analog layers. The number of layers in our models ranged from 0 or 3 convolutional layers and 1 to 3 linear fully connected layers. In these models, we varied the precision of the inputs and weights, the activation functions, and the standard deviation of Gaussian noise. **Table 1** summarizes the parameter space explored in our work (see Appendix for hyperparameter definitions).

| *Hyperparameter* | *Parameters Tested* |
|---:|---|
| *Convolutional Layers* | 0, or 3 layers with kernel size 3x3 |
| *Linear Layers* | 1, 2, or 3 layers |
| *Activations* | Identity, rectified linear unit (ReLU), LeakyReLU, Tanh, exponential linear unit (ELU), sigmoid linear unit/swish function (SiLU), or gaussian error linear unit (GeLU) |
| *Normalization* | None, Clamp (±1), $L^1$Norm, $L^2$Norm, $L^1$NormW, $L^2$NormW, $L^1$NormM, $L^2$NormM, $L^1$NormWM, and $L^2$NormWM |
| *Precision Class* | Reduce Precision, or Stochastic Reduce Precision Layer |
| *Bit Precision* | 2, 4, or 6-bits |
| *Noise Class* | Gaussian Noise |
| *EP* | 0.25, 0.50, or 0.75 ($\sigma \in [0.006, 0.523]$) |
| *Dataset* | MNIST, Fashion MNIST, or CIFAR-10 |
| *Optimizer* | Adam |
| *Optimizer parameters* | Learning rate = 0.001, betas = (0.9, 0.999), weight decay = 0 |
| *Loss* | Cross Entropy Loss |
| *Epochs* | 10 |
| *Batch Size* | 128 |
| *Training to Testing ratio* | 80:20 |

**Table 1:** Table summarizing the hyperparameters explored in Small-Scale Analog Neural Networks. Definitions for Normalization, Noise, and Reduce Precision layers as well as error probability (EP) are detailed in the Appendix.

To more efficiently determine the optimal hyperparameters for our analog models, we employed a simple elimination process: hyperparameters that rarely or never achieve high accuracy are filtered out and no longer tested with other hyperparameter combinations. For example, we found that no model using the $L^1$Norm function in the normalization layers was able to achieve accuracy better than Clamp on any dataset and thus was eliminated from subsequent models. To establish a baseline for digital model accuracy, we begin with an analysis of full precision models with different layer types and hyperparameter combinations (i.e., various normalization and activation classes) as shown in **Table 2**. These top-performing models all have inputs and weights resolution of 16-bits and no added noise, demonstrating the maximum performance that can be expected from our various hyperparameter combinations for the three datasets used for training and testing. As one would expect, we see that the performance of the networks improves with an increasing number of layers and for the more complex CIFAR-10 dataset, using convolutional layers is necessary to achieve >30% accuracy. It is worth noting that convolutional layers do not improve test accuracy for MNIST or FashionMNIST datasets provided there are at least 3 linear layers. With this baseline established, we then explored role of activation function for the classification models in **Table 2** with experimentally relevant precision (6-bits or less for both inputs and weights). These results shown in **Figure 4a** indicate the choice of activation does not strongly influence the performance of small neural networks. This is especially the case for simple datasets which are easily separable such as MNIST and FashionMNIST. However, we do find that optimizing the nonlinear activation function is more important for complex models with a larger depth. We further explore the influence of activation class on the accuracy of a more complex, 9-layer CIFAR-10 model and discuss why certain activations may be preferable for improved performance in the following subsection. In addition to activation functions, we also explored various normalization classes. From the results in **Figure 4b-c**, Clamp is the only normalization function that performs the best when used as input normalization and weight normalization class. Therefore, Clamp is the only normalization class that is viable to use across the full model in these circumstances. We attribute this to the simplest nature of the clamp which mostly preserves the input and weight characteristics if they are in the range of $[-1, 1]$.

|  | *Maximum Test Accuracy (%)* | | |
|---|---|---|---|
| *Model Layers* | **MNIST** | **FashionMNIST** | **CIFAR-10** |
| *1 Linear* | 92.75 | 84.39 | 30.40 |
| *2 Linear* | 97.74 | 88.15 | 38.80 |
| *3 Linear* | 98.10 | 88.89 | 41.31 |
| *3 Conv. + 1 Linear* | 98.91 | 88.73 | 65.07 |
| *3 Conv. + 2 Linear* | 98.91 | 89.42 | 67.34 |
| *3 Conv. + 3 Linear* | 99.06 | 89.07 | 67.85 |

**Table 2:** Maximum test accuracy at full precision (16-bit floating-point numbers used for weights and inputs) for various numbers of convolutional and linear layers. Models trained on the CIFAR-10 dataset require convolutional layers to achieve a reasonable accuracy while the simpler MNIST and FashionMNIST datasets can reach high classification accuracy with only linear layers.

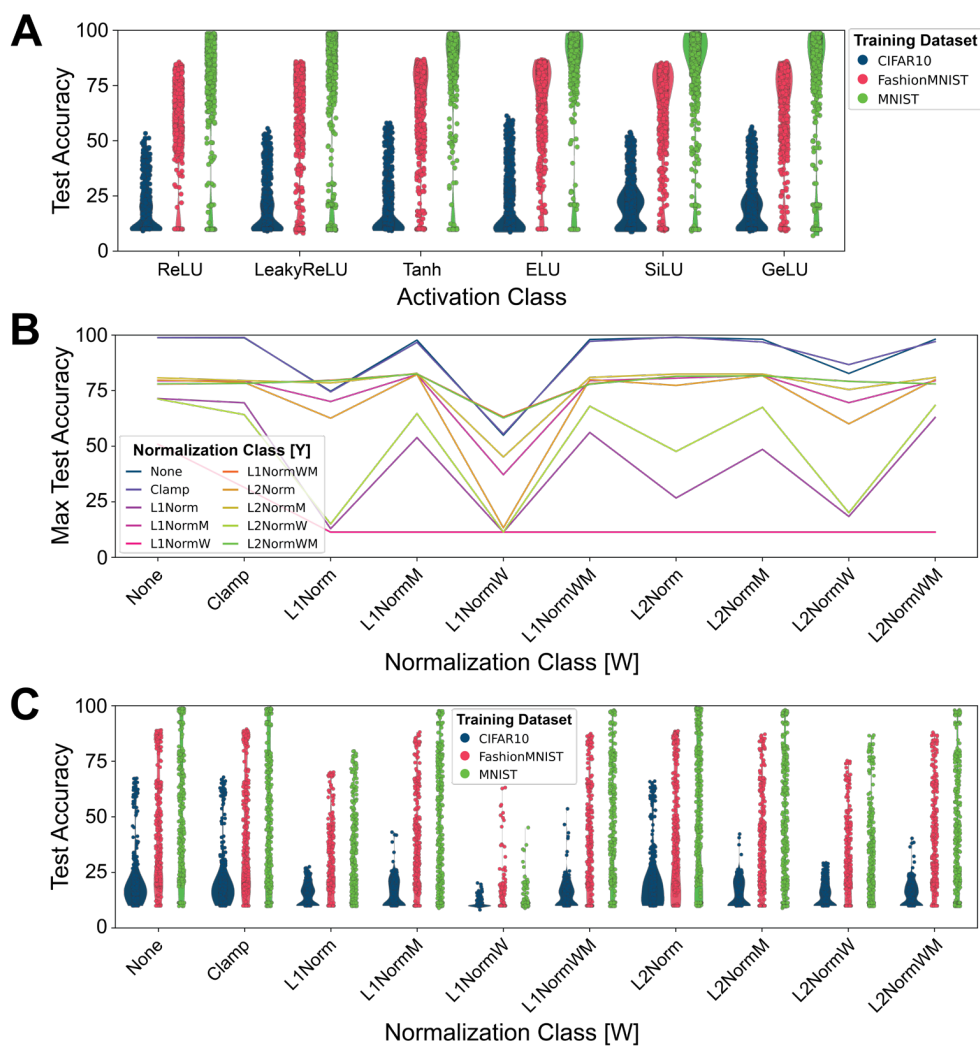

**Figure 4: Test accuracy for various hyperparameter combinations without added noise layers. a)** Scatter plots of the test accuracy for limited precision analog models (i.e., the precision of weights and inputs were 6-bit or less) as a function of activation class used. All activations are able to achieve similar performance. **b)** Influence of normalization on the inputs [Y] and weights [W] on the test accuracy for full precision (16-bit floating-point) models. **c)** Scatter plots of the test accuracy for full precision model as a function of normalization class used for the weights [W].

Having performed an initial analysis on the influence of activation and normalization functions within our analog photonic neural networks, we move on to explore the effects of limited precision and noise. **Figure 5a** shows that both the Reduce Precision (RP) and Stochastic Reduce Precision (SRP) classes perform well, but we also see that SRP is significantly more robust (i.e., the test accuracy of a model with stochastic rounding is higher on average) than RP for models with very low (2-bit) precision. **Figure 5b-c** shows the average and maximum test accuracy of the models explored as a function of the bit precision of the layer inputs [Y] and weights [W]. As expected, we see that as the bit precision increases, the maximum test accuracy of the model also increases. A more detailed look at the results of **Figure 5c** provides some interesting observations which can provide guidelines for the design of analog photonic neural network accelerators. First, we see that increasing the precision of the weights has a larger impact on the overall accuracy of the model than the same increase in the input precision. For example, we see a ~2× improvement for the CIFAR-10 dataset when the precision of the weights is doubled from 2-bits to 4-bits compared to a ~1.4× improvement when the precision of the inputs is increased from 2-bits to 4-bits. Secondly, the amount of precision needed for the inputs and weights is largely dependent on the complexity of the data. From **Figure 5c**, we see that for the MNIST dataset, a precision of 4-bits for the weights and 2-bits for the inputs gives close to the maximum achievable accuracy for the number of layers used (97.38% compared to a maximum accuracy of 99.06% from **Table 2**). We see the importance of precision for both inputs and weights increase more dramatically for models trained on the FashionMNIST and CIFAR-10 datasets. This makes intuitive sense since the MNIST dataset is binary (i.e., black and white images), while the FashionMNIST and CIFAR-10 use greyscale and 8-bit RGB images, respectively. These observations are helpful to consider when designing photonic hardware since a platform with high precision weights and lower precision inputs (i.e., optical signaling with less modulation accuracy) will likely lead to better performing models than vice versa.

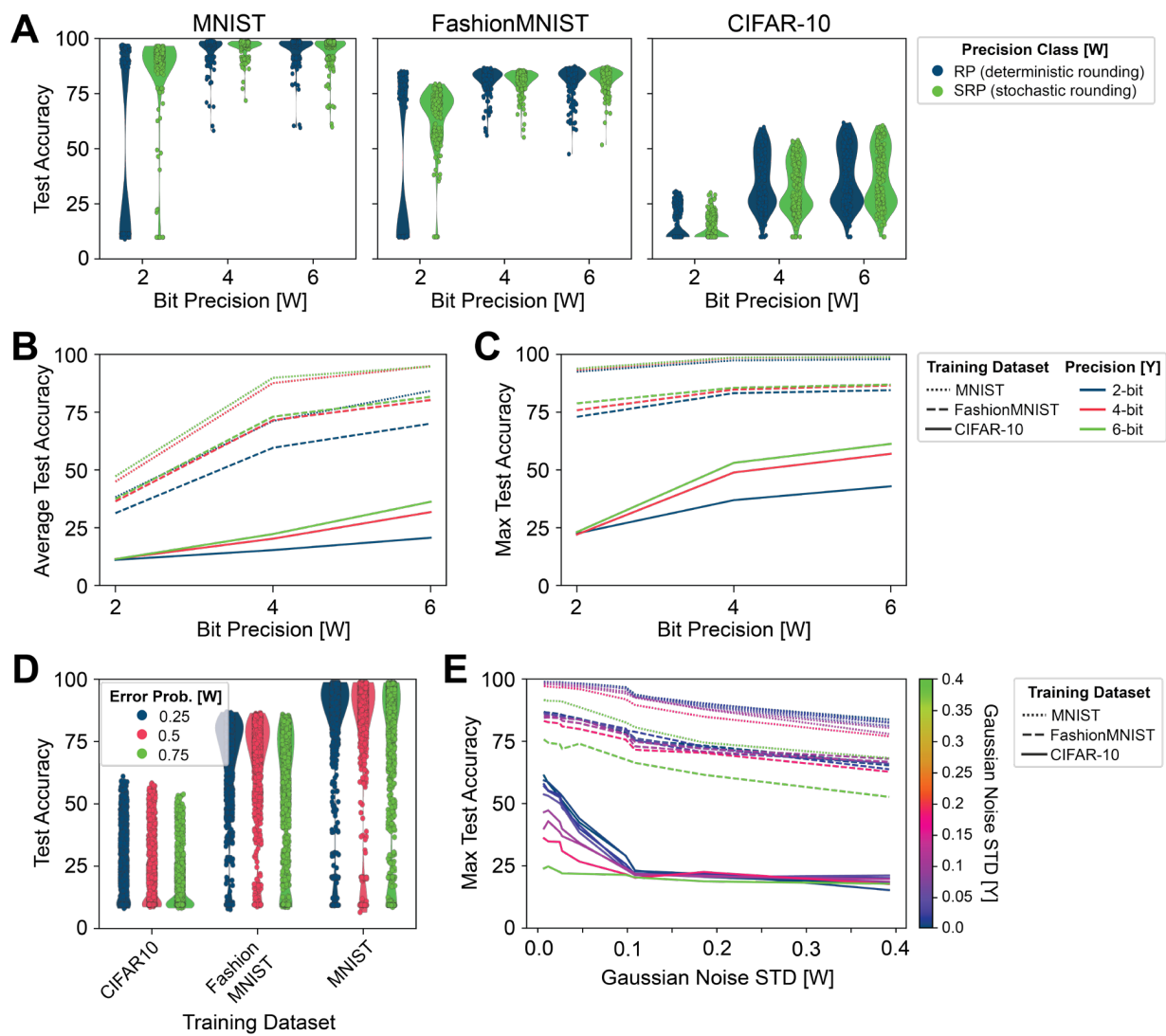

**Figure 5: Influence of reduced precision and added noise on model accuracy. a)** Scatter plots showing the influence of deterministic versus stochastic rounding method for reduce precision layer of the photonic weights. **b)** Average and **c)** maximum test accuracy for three datasets as a function of weight [W] and input [Y] bit precision. The bit precision of weights has a more significant impact than the precision of the inputs and more complex models require higher bit precision for both inputs and weights. **d)** Scatter plot showing the influence of weight error probability (EP) for different datasets. Models trained on simpler data and with higher analog precision are less impacted by random errors. **e)** Impact of the standard deviation of gaussian noise on the maximum test accuracy of the models explored with reduced precision. Like for the case of precision, added noise in the trained weights [W] has a greater impact than noise in the inputs [Y].

In **Figure 5d-e**, we plot the effects of added Gaussian noise on the weights and inputs of our image classification models. To provide an intuitive understanding of the effect of noise, we have mathematically defined the term error probability (or "EP") as the probability that an input value "$x$" will take on a different value "$y$" after passing through a noise layer followed by a reduce precision layer. For example, an EP of 0.75 means that 75% of the input or weight data is a different value than expected after quantization (see Appendix for further details). **Figure 5d** shows a scatter plot of model test accuracy for three different EP values with an expected inverse correlation between EP and accuracy. Additionally, we see that as the complexity of the datasets increases, the maximum accuracy achieved decreases.

The effects of EP are naturally stronger for photonic networks with a lower bit precision since this corresponds to a larger change from the expected quantized value ($\varepsilon_q \geq 2^{1-N_b}$ for $RP(x) \in [-1,1]$ and $\varepsilon_q \geq 2^{-N_b}$ for $RP(x) \in [0,1]$, where $N_b$ is the number of bits and $\varepsilon_q$ is the quantization error). For the case of gaussian noise, the relationship between EP, bit precision ($N_b$), and standard deviation ($\sigma$) is:

$$\text{EP} = 1 - \text{erf}\left(\frac{1}{2\sqrt{2} * \sigma * (2^{N_b} - 1)}\right)$$

where erf (x) is the error function. We can use the above equation to compare the effects of noise more directly with typical signal-to-noise ratio (SNR) metrics used in the analog domain since SNR $\propto 1/\sigma$. In **Figure 5e**, we plot the maximum test accuracy achieved from the various models tested as a function of the standard deviation of the noise for both inputs ["Y"] and weights ["W"]. A major conclusion of our analog simulations becomes evident once again: precision and noise of the trained weights affect the accuracy of the network much more than the precision and noise of the layer inputs. Therefore, when implementing a physical photonic neural network, low-noise and high-precision weights are the most likely to act as the limiting factor for device performance—especially for complex datasets. This combination of tests resulted in maximum simulated accuracies of 99.14%, 89.06%, and 58.67% for MNIST, FashionMNIST, and CIFAR10 respectively using 10 training epochs to minimize overhead. We found that for CIFAR10, test accuracies increased to 77% for 50 epochs, and can further be increased by using a deeper network (see following section) or by using a different loss function.

### *Applying AnalogVNN to Large Analog Neural Network Models (9-layer)*

After exploring a large hyperparameter space with small 1- to 6-layer linear and convolutional neural networks, we trained a larger 9-layer convolutional neural network with ~1.7 million model parameters [39] using AnalogVNN to see if our conclusions still hold for more complex analog neural networks (model shown in **Figure 6a and Table 3**). Initial testing performed after the conversion of the model to its analog counterpart was unable to achieve a maximum test accuracy above 70% compared to the maximum test accuracy of 86.24±0.19% achieved by the original full precision (32-bit, floating-point numbers) digital model [8] after training. For this reason, a few additional hyperparameters were tested including batch size, color/grayscale, batch normalization, gradient functions, optimizer, and learning rate (see **Table 3** for a full list of hyperparameters explored). It was found that batch size plays an important role in training the 9-layer model, as is shown in **Figure 6b**. We attribute this strong dependence on batch size to the generalization gap arising from pre-normalized data provided by PyTorch [57], [58]. Due to the computation constraints, a batch size of 512 was chosen for the remainder of the training.

In **Figure 6b** we also observe that three specific activations (GeLU, SiLU, and LeakyReLU) provide the best model performance on average and are less impacted by batch size than the other activation functions tested [23], [59]. We hypothesize that the poorer performance of "Tanh" and "ELU" functions is related to near constant slope near zero, thus causing instabilities during training when the output of a layer is close to zero and noise is present. We also suspect that the "ReLU" function behaves poorly compared to the similar GeLU and LeakyReLU since it has a zero slope in the negative domain and does not provide negative feedback between layers [23], [59]. A non-zero slope in the negative domain is important when noise is present since for values close to zero, random noise can cause small-valued positive numbers to become negative and vice versa. Using a ReLU activation will cause these small negative values to be zero which makes training challenging. These observations have a direct impact on the design of photonic hardware since negative optical signaling between layers (i.e., what we have defined as "inputs") is only allowed for coherent photonic platforms with both amplitude and phase control. Matrix-vector operations containing negative-valued vector elements will require two separate matrix operations for incoherent architectures:

$$\boldsymbol{W}Y^+ - \boldsymbol{W}Y^- = B, Y^+ = \max(0, Y) \text{ and } Y^- = \min(0, Y)$$

$$Y^+, Y^- \in [0,1] \text{ and } Y, B, \boldsymbol{W} \in [-1,1]$$

where $\boldsymbol{W}$ is the weight matrix, $Y^+$ and $Y^-$ are the sub-vectors containing positive and negative elements of the input vector $Y$ respectively, and $B$ is the resulting output vector. Thus, for activation functions with negative output values, incoherent architectures will either require duplicate hardware or twice the processing time to perform these negative matrix operations. This makes the development of training algorithms which use positive-valued activations (like ReLU or sigmoid) highly attractive. It is also worth noting that many of the experimental demonstrations and modeling of photonic neural networks have been limited to small neural networks (3-layers or less) and on simplistic datasets (MNIST or FashionMNIST). Our initial results show that for larger neural networks trained on more complex (and arguably more useful) datasets, assumptions made using simpler models may not be generalizable to more complex models.

| *Hyperparameter* | *Parameters Tested* |
|---:|---|
| *Convolutional Layers* | 6 layers with kernel size 3x3 |
| *Linear Layers* | 3 layers |
| *Activations* | Identity, rectified linear unit (ReLU), LeakyReLU, Tanh, exponential linear unit (ELU), sigmoid linear unit/swish function (SiLU), or gaussian error linear unit (GeLU) |
| *Normalization* | None, Clamp (±1), $L^1$Norm, $L^2$Norm, $L^1$NormW, $L^2$NormW, $L^1$NormM, $L^2$NormM, $L^1$NormWM, and $L^2$NormWM |
| *Precision Class* | Reduce Precision, or Stochastic Reduce Precision Layer |
| *Bit Precision* | 2, 3, 4, 5 or 6-bits |
| *Noise Class* | Gaussian |
| *EP* | 0.20, 0.40, 0.60, or 0.80 ($\sigma \in [0.006, 0.657]$) |
| *Optimizer* | Adam |
| *Optimizer parameters* | Learning rate $= 0.001$, betas $= (0.9, 0.999)$, weight decay $= 0$ |
| *Loss* | Cross Entropy Loss |
| *Dataset* | CIFAR-10 |
| *Epochs* | 200 |
| *Batch Size* | 128, 256, 384, or 512 |
| *Training to Testing ratio* | 80:20 |

**Table 3:** Table summarizing the hyperparameters explored for the 9-layer CIFAR-10 classification models.

Also of interest, is how normalization of the weight matrix effects the classification accuracy of deeper analog models. When compared to the results of **Figure 4b**, **Figure 6c** shows that a deeper model is much more sensitive to the normalization class than a shallow model. Both Clamp and L$^2$Norm performed well for the weights, but only the Clamp normalization class was

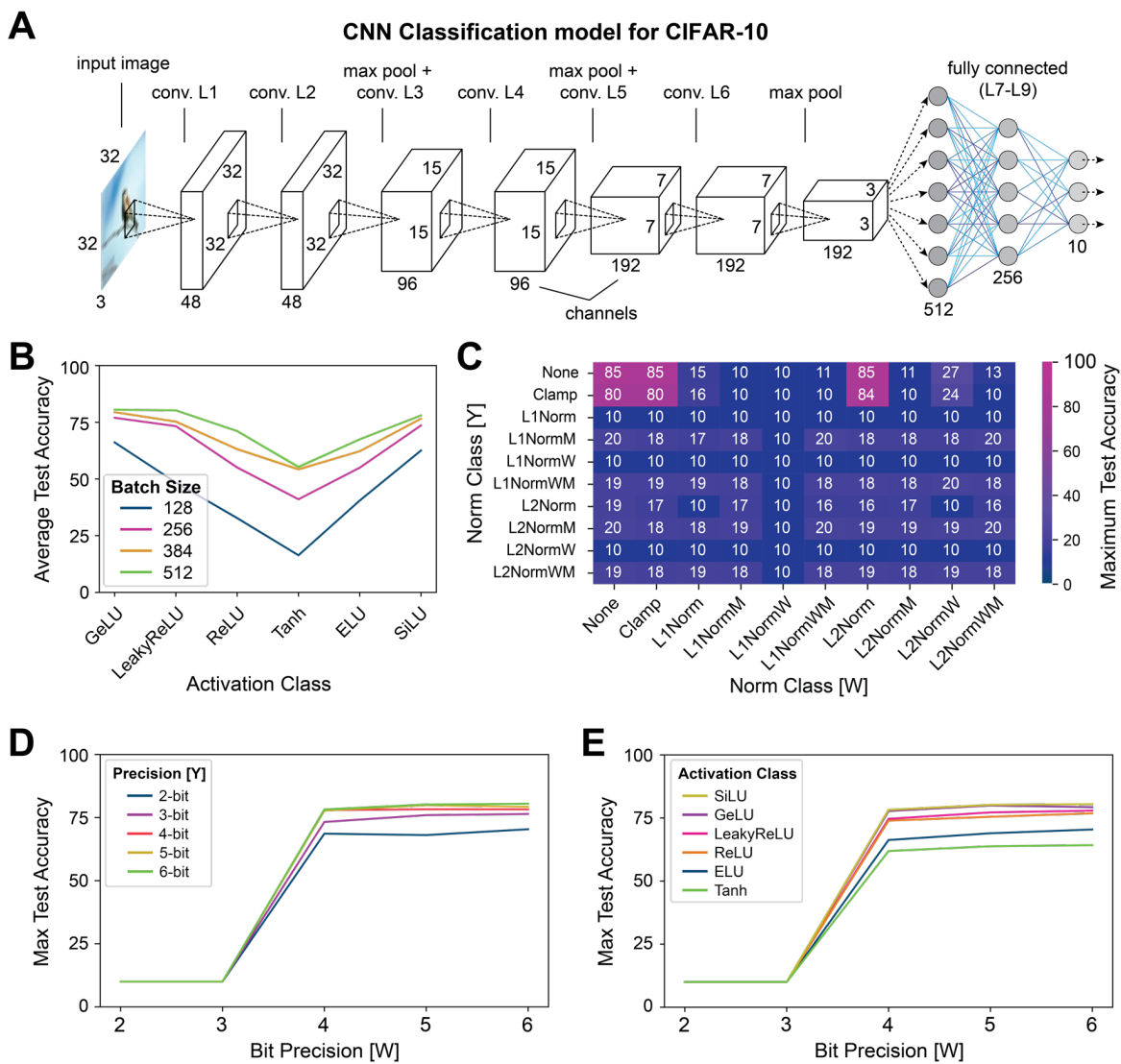

**Figure 6: Modeling a 9-layer CNN using AnalogVNN without added noise. a)** Overview of the 9-layer CNN model used for classifying the CIFAR-10 dataset. **b)** Average test accuracy as a function of batch size and activation class. A batch size of 512 was used to achieve high accuracy for 200 training epochs used throughout the rest of our studies. **c)** Maximum test accuracy of the models for different weight [W] and input [Y] normalization classes. **d)** Impact of input [Y] and weight [W] precision on maximum test accuracy. **e)** and the effect of activation functions for inputs with 6-bit precision. For learning, a minimum of 4-bits precision are needed for the weights. **Note:** full precision for weights and inputs was used in **b)** and **c)** while reduced precision was used in **d)** and **e)**.

able to achieve high accuracies when applied to the input vectors between layers. We therefore chose to limit further tests to Clamp normalization layers for both inputs and weights as Clamp is the simplest normalization function to physically implement in hardware. A critical observation can be seen in **Figure 6d** where we show that a minimum weight precision of 4 bits is required to achieve any learning in more complex datasets like CIFAR-10. As in the smaller models in the previous section, we see here that input precision is less important than weight precision. **Figure 6e** shows the influence of activation as a function weight precision (6-bit precision was used for the inputs in these tests). As in **Figure 6b**, we see SiLU and GeLU result in the highest test accuracy for models with reduced precision. [23], [59]

In a final study, we simulate the impact of added noise on the model accuracy. We also explore the importance of retaining color for the CIFAR-10 dataset and find a minor impact on the resulting test accuracy of our models (see **Figure 7a**). However, this conclusion may not be true for models trained on more complex image datasets such as ImageNet [60]. In **Figure 7b**, we plot the maximum test accuracy of different activation functions in the presence of noise added to the weights (layer inputs were 6-bit with EP = 0.2). While initially similar to the results in **Figure 6b** for low EP, after the introduction of both noise and the reduced precision we see that GeLU and SiLU performed much better than LeakyReLU and other activations for higher weight error. This highlights the importance of using activations with a constant slope near zero for both positive and negative values during training [23], [59]. As observed in previous results, increasing bit precision or decreasing noise of the weights has the greatest impact on the maximum test accuracy the model can achieve (**Figure 7b-c**). The correlation of these two effects—bit precision and noise in weights—is shown in **Figure 7c**. To more directly compare with typical SNR metrics used for signals in the analog domain, we plot the maximum test accuracy achieved from the various models tested as a function of the standard deviation of the noise for both inputs ($Y$) and weights ($W$) in **Figure 7d**. Our major conclusion from previous tests on smaller models still holds true: the precision and noise of the weights affect the accuracy of the network much more than the precision and noise of the inputs. This is clearly seen in both **Figure 5e** for small models and **Figure 7d** for larger models, particularly as the complexity of the dataset increases.

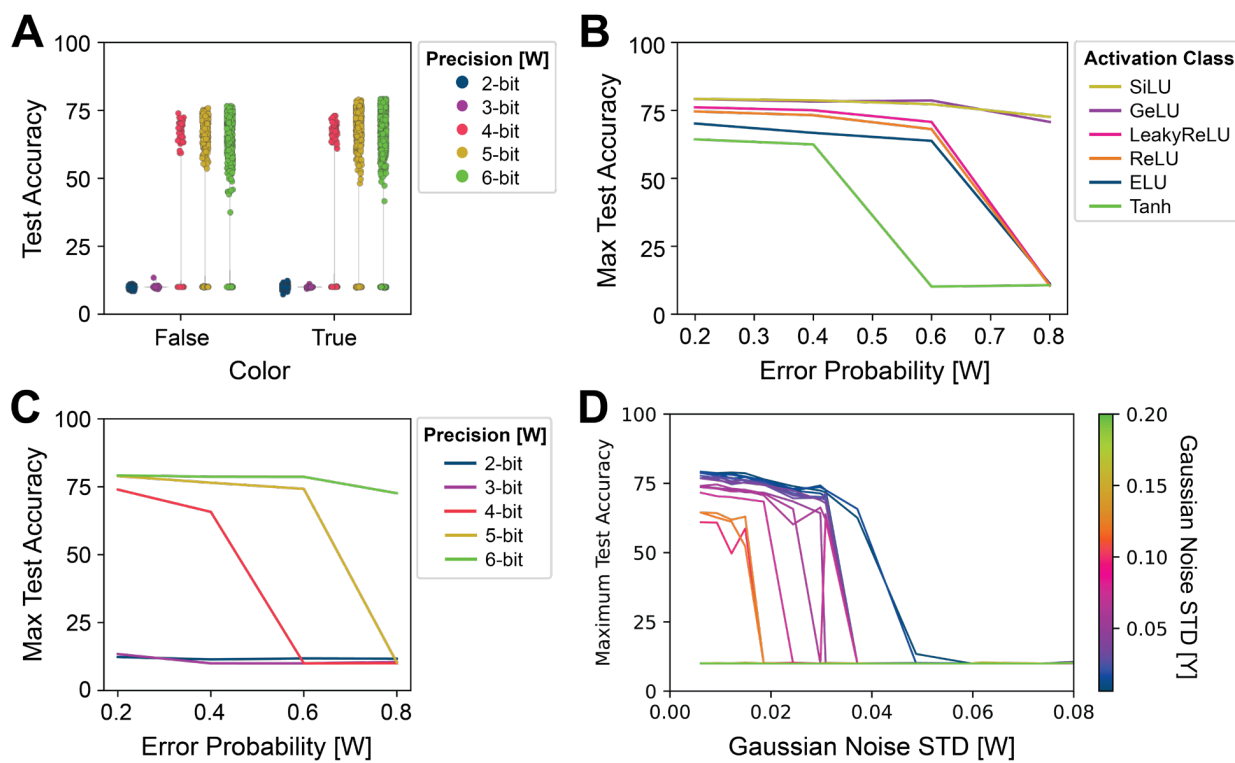

**Figure 7: Modeling a 9-layer CNN using AnalogVNN with both reduce precision and added noise. a)** Removing image color has a minimum impact on the training accuracy of our models. **b)** Impact of activation functions on models showing best model performance from GeLU, and SiLU for all the levels of noise. Maximum test accuracy as a function of weight [W] and input [Y]. **c)** The correlation of precision and error probability for weights. **d)** Impact of the standard deviation of gaussian noise on test accuracy.

## V. Limitations to AnalogVNN

Our layer-based analog modeling approach shown in **Figure 1** works well for typical feedforward neural networks comprised of linear and convolutional layers but is not applicable to all neural network types or photonic hardware implementations. For example, more complex neural models with feedback or dynamic response (e.g., LSTMs [61] and recurrent neural networks [62]) would require additional adjustments for a proper approximation which is currently outside the scope of this work. Additionally, we note that while free-space approaches which use optical interference from diffractive or scattering media can be approximated with complex matrix operations [63], [64], the mapping between trained weights from an equivalent AnalogVNN model is non-trivial and would require additional functionality not yet implemented. Despite these limitations, AnalogVNN can be used to simulate the majority of photonic AI hardware accelerators which directly implement matrix-vector operations.

## VI. Conclusion

Using AnalogVNN, we were able to model over 570,000 unique hyperparameter combinations for 1- to 6-layer photonic analog image classifiers trained on three different datasets. Guided by these results for small networks, we then successfully optimized various hyperparameters for an analog 9-layer CIFAR-10 convolutional neural network with ~1.7 million trainable parameters [8], [39]. Our simulation of noisy, limited precision analog neural networks provides us with a better understanding of the minimum requirements and considerations that are needed for photonic neural network design. We found that simple 2-to-6-bit on-chip photonic circuits can be used for both small-scale linear and convolutional neural networks when training on simple datasets. However, higher precision (≥4-bits) is required for the larger 9-layer CIFAR-10 models produce appreciable accuracies. This is encouraging for photonic in-memory computing approaches, particularly those utilizing phase-change materials as photonic memory cells which have demonstrated up to 6-bits of precision per memory cell [49]. Our simulations revealed that for both small and large models, classification accuracies are more strongly dependent on bit precision and noise of the weights rather than the precision and noise of the inputs. We also observed that differentiable activation functions like GeLU and SiLU performed best at higher levels of noise, raising the importance of encoding both positive and negative numbers in optical input signals between layers [23], [59]. In summary, we have shown that AnalogVNN can be used to simulate photonic neural networks and help significantly narrow down the analog design space by eliminating non-optimal hyperparameters. This approach can significantly reduce valuable time and cost for photonic analog hardware design and testing. Due to its generality and flexibility in layer definitions, we also expect that AnalogVNN will be a useful framework to simulate analog hardware in other domains such as electronic, magnetic, or spintronic systems [8], [9], [28], [29].

## VII. Acknowledgements

This work was supported in part by the U.S. National Science Foundation under Grants ECCS-2028624, DMR-2003325, and CISE-2105972. N.Y. acknowledges support from the University of Pittsburgh Momentum Fund.

This research was also supported in part by the University of Pittsburgh Center for Research Computing, RRID:SCR_022735, through the resources provided. Specifically, this work used the H2P cluster, which is supported by NSF award number OAC-2117681.

# Appendix I: Definition of Analog Layers

To simulate an analog photonic neural network, each layer is constrained by reduced precision, added noise, and signal normalization which requires computation in addition to the standard digital operations (such as multiply-accumulate and nonlinear operations). For AnalogVNN, we have found that introducing reduce precision, noise, and normalization layers in PyTorch results in an increase ranging from 10% to 30% in the overall training and testing time, which is acceptable. However, if we were to modify the multiply-accumulate operation in PyTorch or TensorFlow to include analog effects, the resulting training and testing times would be extremely long. Since there is a need to test thousands of different hyperparameter combinations, this becomes quickly unfeasible for current computation technology. In the following sections, we describe the inner workings of Reduce Precision, Noise, and Normalization layers.

## I. Normalization Layers

In the analog domain, there is an upper limit to the maximum analog signal which can be generated (for photonics, it can be the laser power or the modulation amplitude). To maximize the compatibility of algorithms and encoding in the analog domain between various hardware platforms, it is logical to represent inputs and weights as values in the range $[-1,1]$ such that the result of real-valued linear operations (e.g., matrix-vector multiplications) can simply be scaled by a factor. This normalization is also common in neural networks to reduce exploding and vanishing gradients during training [65], [66]. To represent this in our simulated model, we use Normalization layers in which ±1 represents the maximum of the signal (differing by a $\pi$ phase shift in the case of coherent architectures) and 0 represents no optical signal. This can be done using the following layers:

### A. LpNormW

$L^p$NormW applies p-normalization to the input batch matrix or weight matrix [23]. This process normalizes the entire matrix by the same scalar value. $L^p$NormW is defined as follows:

$$L^p\text{NormW}(x) = \frac{x}{\|x\|_p} = \frac{x}{\sqrt[p]{\sum |x|^p}}$$

$$L^p\text{NormWM}(x) = \frac{L^p\text{NormW}(x)}{max(|L^p\text{ NormW}(x)|)}$$

where:
  $x$ is the input matrix.
  $p$ is a positive integer.

### B. $L^p$Norm

$L^p$Norm normalizes by dividing each element of the matrix $x_{ij\ldots k}$ by p-normalization of $x_{j\ldots k}$ [23]. That is, it normalizes each image in the input batch, each row in the weight matrix of the

linear layer, and each output channel of the weight matrix of the convolution layer. $L^p$Norm is defined as:

$$L^p\text{Norm}(x) = \left[ x_{ij...k} \rightarrow \frac{x_{ij...k}}{\sqrt[p]{\sum_{j...k} |x_{ij...k}|^p}} \right]$$

$$L^p\text{NormM}(x) = \frac{L^p\text{Norm}(x)}{max(|L^p\text{Norm}(x)|)}$$

where:
  $x$ is the input weight matrix.
  $i, j...k$ are indexes of the matrix.
  $p$ is a positive integer.

## C. Clamp$_{pq}$

*Clamp$_{pq}$* cuts the signal off at the upper and lower limits by applying a minimum value of p and a maximum value of q to the input signal or weight matrix [23].

$$\text{Clamp}_{pq}(x) = \begin{cases} q & \text{if } q < x, \\ x & \text{if } p \leq x \leq q, \\ p & \text{if } p > x \end{cases}$$

where:
  $p, q \in \Re$ ($p \leq q$, Default value for photonics $p = -1$ and $q = 1$).

## II. Reduce Precision Layers

To explore the impact of representing high precision, digital numbers in analog photonic neural networks with limited precision, we have implemented two reduced precision layer types. The first type, *Reduce Precision*, is a simple round-to-nearest transformation where all high precision, 32-bit input values are deterministically mapped to a corresponding value of a fixed precision range. The second type, *Stochastic Reduce Precision*, uses stochastic rounding to limit the precision, which adds a probabilistic weighting to the rounding process. These have been implemented by modifying the equations from Gupta et al. [67] as discussed below:

### A. Reduce Precision (RP)

Reduce Precision layer applies a round-to-nearest transformation to the input based on the precision (number of discrete levels) and divide parameter (the threshold for rounding between neighboring discrete levels) [8], [67]. The relationship between the parameters and discrete levels is illustrated in **Figure 8a-b**.

$$RP(x) = \text{sign}(x*p) * \max(\lfloor |x*p| \rfloor, \lceil |x*p| - d \rceil) * \frac{1}{p}$$

$$\text{Step Width} = \frac{1}{p}$$

$$R_{RP}(a,b) = \{x \in [a,b] \mid RP(x) = x\}$$

$$\text{size}(R_{RP}(a,b)) = (RP(b) - RP(a)) * p + 1$$

where:

$x$ is the original number in full precision.

$p$ is the analog precision of the input signal, output signal, or weights (p ∈ Natural Numbers, $number\ of\ bits = \log_2(p+1)$).

$d$ is the divide parameter ($0 \leq d \leq 1$, *default value* = 0.5) which determines whether $x$ is rounded to a discrete level higher or lower than the original value

**Step Width** is the range of $x$ for which $RP(x)$ gives the same value (also the minimum separation between $p$ discrete levels).

***R<sub>RP</sub>*** is the range of values Reduce Precision can take between $a$ and $b$ ($a < b$).

## B. Stochastic Reduce Precision (SRP)

Stochastic Reduce Precision layer applies a stochastic rounding transformation to the input based on the precision parameters [8], [67]. Thus, the input $x$ will be stochastically rounded to a lower or higher discrete value with a probability proportional to its proximity to that discrete value as shown in **Figure 8c**. This can improve training stability as we discuss in more detail (see *Results and Physical Implementation* section).

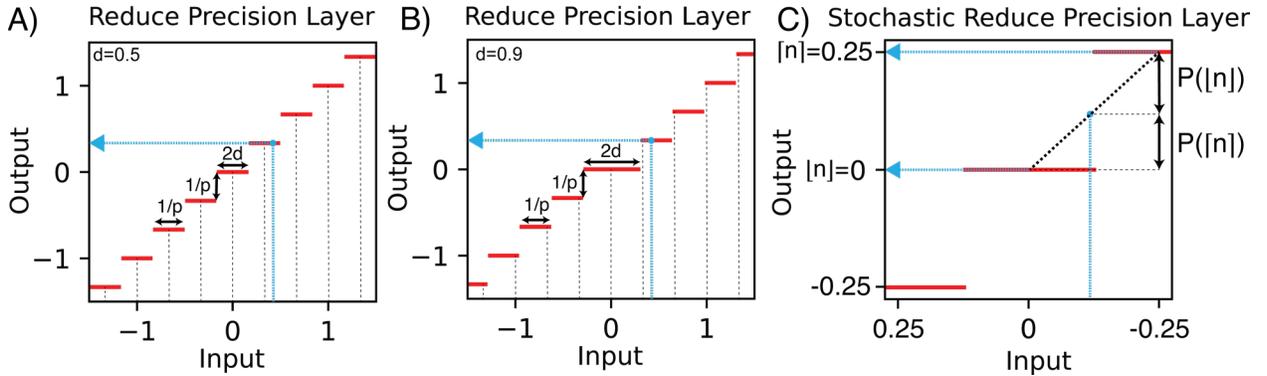

**Figure 8.** Illustration of rounding in Reduce Precision and Stochastic Reduce Precision layers.

$$SRP(x) = \text{sign}(x*p) * f(|x*p|) * \frac{1}{p}$$

$$f(x) = \begin{cases} \lfloor x \rfloor & \text{if } r \leq 1 - |\lfloor x \rfloor - x|, \\ \lceil x \rceil & \text{otherwise,} \end{cases}$$

where:
> $r$ is a uniformly distributed random number between 0 and 1.
> $p$ is the analog precision ($p \in$ Natural Numbers, $number\ of\ bits = \log_2(p+1)$).
> $f(x)$ is the stochastic rounding function.

## III. Noise Layers

### A. Error Probability (EP)

We have defined an information loss parameter, "error probability" or "EP," as the probability that a reduced precision digital value (e.g., "1011") will acquire a different digital value (e.g., "1010" or "1100") after passing through the noise layer (i.e., the probability that the digital values transmitted and detected are different after passing through the analog channel). This is a similar concept to the bit error ratio (BER) used in digital communications, but for numbers with multiple bits of resolution. While SNR (signal-to-noise ratio) is inversely proportional to $\sigma$, the standard deviation of the signal noise, EP is indirectly proportional to $\sigma$. However, we choose EP since it provides a more intuitive understanding of the effect of noise in an analog system from a digital perspective. It is also similar to the rate parameter used in PyTorch's Dropout Layer [23], though different in function. EP is defined as follows:

$$EP = 1 - \frac{\int_{q=a}^{b} \int_{p=-\infty}^{\infty} \text{sign}\left(\delta(RP(p) - RP(q))\right) * PDF_{N_{RP(q)}}(p)\ dp\ dq}{|b - a|}$$

$$EP = 1 - \frac{\int_{q=a}^{b} \int_{p=\max\left(RP(q) - \frac{S}{2}, a\right)}^{\min\left(RP(q) + \frac{S}{2}, b\right)} PDF_{N_{RP(q)}}(p)\ dp\ dq}{|b - a|}$$

$$EP = 1 - \frac{1}{\text{size}(R_{RP}(a,b)) - 1} * \sum_{q \in S_{RP}(a,b)} \int_{\max\left(q - \frac{S}{2}, a\right)}^{\min\left(q + \frac{S}{2}, b\right)} PDF_{N_q}(p)\ dp$$

$$EP = 1 - \frac{1}{\text{size}(R_{RP}(a,b)) - 1} * \sum_{q \in S_{RP}(a,b)} \left[CDF_{N_q}(p)\right]_{\max\left(q - \frac{S}{2}, a\right)}^{\min\left(q + \frac{S}{2}, b\right)}$$

For noise distributions invariant to linear transformations (e.g., Uniform, Normal, Laplace, etc.), the EP equation is as follows:

$$EP = 2 * CDF_{N_0}\left(-\frac{1}{2 * (2^{bit\_precision} - 1)}\right)$$

For Gaussian Noise:

$$EP = 1 - \text{erf}\left(\frac{1}{2\sqrt{2} * \sigma * (2^{bit\_precision} - 1)}\right)$$

where:

**EP** is in the range [0, 1].
**δ** is the Dirac Delta function.
**RP** is the Reduce Precision function (for the above equation, $d = 0.5$).
**s** is the step width of reduce precision function.
**S$_{RP}$** (a, b) is $\{x \in [a, b] \mid RP(x) = x\}$.
**PDF$_x$** is the probability density function for the noise distribution, x.
**CDF$_x$** is the cumulative density function for the noise distribution, x.
**N$_x$** is the noise function around point x. (for Gaussian Noise, x = *mean* and for Poisson Noise, x = *rate*).
**a, b** are the limits of the analog signal domain (for photonics $a = -1$ and $b = 1$).

## B. Gaussian Noise Layer

Gaussian noise can be added to the input with the noise distribution scaled according to the analog precision and the EP value (illustrated in **Figure 9**). Depending on the EP value and the neural model, gaussian noise can behave like a regularization layer, increasing robustness during training [68], [69]. The relationship between the standard deviation of gaussian noise, EP, and precision is given as follows:

$$\text{EP} = 1 - \text{erf}\left(\frac{1}{2\sqrt{2} * \sigma * (2^{\text{bit\_precision}} - 1)}\right)$$

$$\sigma = \frac{1}{2\sqrt{2} * (2^{\text{bit\_precision}} - 1) * \text{erf}^{-1}(1 - \text{EP})}$$

where:
**σ** is the standard deviation of Gaussian Noise.
**EP** is the error probability ($0 \leq EP \leq 1$).
erf is the Gauss Error Function.

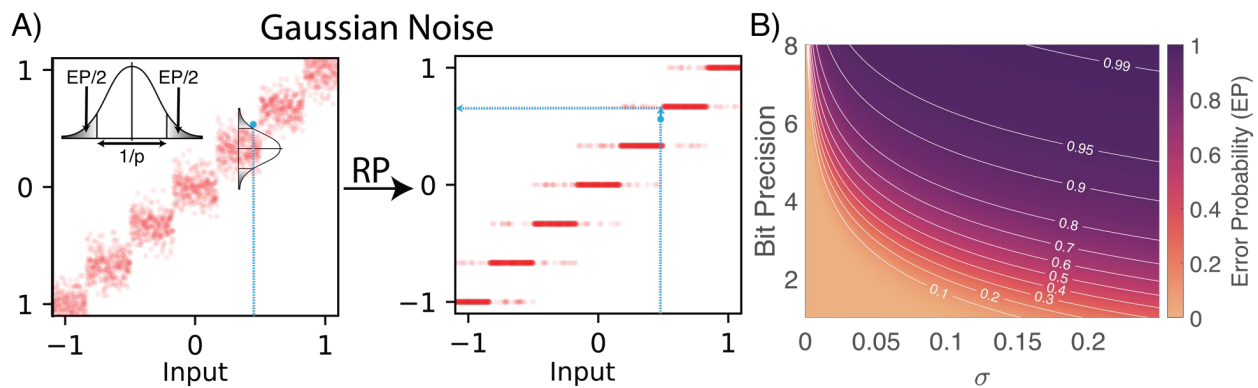

**Figure 9. a)** Example of Gaussian noise being added to a reduced precision layer. The standard deviation of the noise distribution ($\sigma$) causes a fraction of the reduced precision values (*EP*) to be read out at either a higher or lower bit value once quantized at the output. **b)** Relationship between bit precision, standard deviation, and error probability for a Gaussian noise distribution.